\documentclass[conference]{IEEEtran}
\IEEEoverridecommandlockouts
\usepackage{cite}
\usepackage{amsmath,amssymb,amsfonts}
\usepackage{algorithmic}
\usepackage{graphicx}
\usepackage{textcomp}
\def\BibTeX{{\rm B\kern-.05em{\sc i\kern-.025em b}\kern-.08em
    T\kern-.1667em\lower.7ex\hbox{E}\kern-.125emX}}
\begin{document}

\title{Fast Convergence for Stochastic and Distributed Gradient Descent in the Interpolation Limit\\
}

\author{\IEEEauthorblockN{Partha P Mitra}
\IEEEauthorblockA{\textit{Cold Spring Harbor Laboratory} \\
Cold Spring Harbor, NY11724, USA \\}
}

\maketitle

\begin{abstract}
Modern supervised learning techniques, particularly those using deep nets, involve fitting high dimensional labelled data sets with functions containing very large numbers of parameters. Much of this work is empirical. Interesting phenomena have been observed that require theoretical explanations; however the non-convexity of the loss functions complicates the analysis. Recently it has been proposed that the success of these techniques rests partly in the effectiveness of the simple stochastic gradient descent algorithm in the so called interpolation limit in which all labels are fit perfectly. This analysis is made possible since the SGD algorithm reduces to a stochastic linear system near the interpolating minimum of the loss function. Here we exploit this insight by presenting and analyzing a new distributed algorithm for gradient descent, also in the interpolating limit. 

The distributed SGD algorithm presented in the paper corresponds to gradient descent applied to a simple penalized distributed loss function, $L({\bf w}_1,...,{\bf w}_n) = \Sigma_i l_i({\bf w}_i) + \mu \sum_{<i,j>}|{\bf w}_i-{\bf w}_j|^2$. Here each node holds only one sample, and its own parameter vector. The notation $<i,j>$ denotes edges of a connected graph defining the communication links between nodes. It is shown that this distributed algorithm converges linearly (ie the error reduces exponentially with iteration number), with a rate $1-\frac{\eta}{n}\lambda_{min}(H)<R<1$ where $\lambda_{min}(H)$ is the smallest nonzero eigenvalue of the sample covariance or the Hessian H. In contrast with previous usage of similar penalty functions to enforce consensus between nodes, in the interpolating limit it is not required to take the penalty parameter to infinity for consensus to occur. The analysis further reinforces the utility of the interpolation limit in the theoretical treatment of modern machine learning algorithms. 

\end{abstract}

\begin{IEEEkeywords}
Interpolating limit; Overfitting; Stochastic Gradient Descent; Distributed Gradient Descent. 
\end{IEEEkeywords}

\section{Introduction}
Empirical performance advances in using so-called deep networks for machine learning\cite{goodfellow2016deep} have given rise to a growing body of theoretical work that tries to explain the success of these methods. While this theoretical area is not yet in a mature state, several interesting observations have been made and ideas put forth. One observation is that overfitting surprisingly does not seem to degrade the generalization performance of deep nets in supervised learning\cite{2016arXiv161103530Z},  \cite{2018arXiv180201396B}. A related phenomenon is that relatively simple randomized optimization algorithms, particularly stochastic gradient descent (SGD), are quite effective\cite{goyal2017accurate}, despite the complicated non-convex landscapes of the associated loss functions. 

Recently, these phenomena have been connected and it has been shown that in the overfitting or interpolating limit, stochastic gradient descent converges rapidly to an interpolating minimum of the loss function\cite{ma2017power}. In addition, under suitable assumptions, the performance of the SGD algorithm saturates at relatively small mini-batch sizes, raising questions about the role of data parallelism. 

{\it Role of parallelism:} Data and model parallelism plays an important role in the area of deep nets since both data and model sizes can be very large, and resource constraints enforce the need for such parallelism\cite{dean2012large}. Many algorithmic variants of parallelized SGD have been proposed and studied. Most of these variants employ centralized communications, although some fully distributed algorithms have also been studied\cite{jiang2017collaborative}. Much of this work is empirical. While there exist theoretical results concerning convergence, these can be involved and lack intuitive simplicity. 

{\it Analysis in the Interpolating Limit:} A particular benefit of the interpolating limit is that the analysis can be effectively carried out within the analytically tractable reaches of linear systems theory, with the caveat that it is necessary to understand the behavior in the limit of large system sizes as well as the presence of randomness. We exploit this simplicity to gain insight into the role of data correlations in determining the impact of parallelism on SGD, and to study a fully distributed consensus-based algorithm for gradient descent, with one data sample per computational node. 

{\it Algorithm presented in this paper for Distributed GD:} The algorithm corresponds to gradient descent applied to a penalized distributed loss function, 

\begin{equation}
L({\bf w}_1,...,{\bf w}_n) = \Sigma_i l_i({\bf w}_i) + \mu \sum_{<i,j>}|{\bf w}_i-{\bf w}_j|^2
\end{equation}

Here each node holds only {\it one} sample (the logical extreme of data parallelism), and is allowed its own parameter vector. The notation $<i,j>$ denotes edges of a connected graph defining the communication links between nodes. There is no central parameter server: the DGD algorithm following from the above setup only involves communication of the parameter vector estimates between neighboring nodes. At the minimum of the distributed loss function, each node recovers the same weight corresponding to an interpolating minimum of the centralized loss, ${\bf w}_i={\bf w}^*=argmin L({\bf w})$. It is shown that this algorithm converges linearly (ie the approximation error decreases exponentially with iteration). The proposed distributed algorithm points to a number of avenues for future work. Note that variants of this algorithms could assign more than one sample to each compute node, in a deterministic or stochastic manner. 

\section{Relation to previous work}

There is a large relevant literature on stochastic optimization, including considerations of parallelism and distributed computation and spanning multiple disciplines. It is not practical to review this literature here, but we provide a few salient references and pointers. Closely related to the current work is the analysis of the exponential convergence of the randomized Kacmarz algorithm\cite{strohmer2009randomized}, which is analogous to SGD with a minibatch size of 1, and its distributed version\cite{kamath2015distributed}, as well as recent work on consensus-based distributed SGD for learning deep nets\cite{jiang2017collaborative}. However the distributed algorithm obtained by performing gradient descent on the loss function in Eq.(1), does not appear to have been analyzed in the previous literature, particularly in the interpolating limit which greatly simplifies the analysis. 

{\it Relation to ADMM and penalty based approaches:} There is also a relevant body of literature on penalty-based methods for distributed optimization, as well as ADMM related approaches\cite{boyd2011distributed} where the consensus constraint is enforced using Lagrange multipliers. While the current approach is in effect penalty-based, the important point is that {\it in the interpolating limit} the penalty term does {\it not} have to be made large to achieve consensus. The {\it exact} optimum is obtained for any value of the penalty term. Thus the penalty parameter could be optimized to speed up convergence. In contrast with ADMM the algorithm does not require the introduction of dual variables. Nevertheless, one future direction from the current work would be to re-examine these other methods of performing distributed optimization specifically in the interpolating limit, which may lead to analytical simplicities. 

The energy function represented in Eq.(1) has a standard form familiar in statistical physics. It represents a sum of on-site energies and quadratic or elastic couplings between neighboring sites (on a suitable graph). In the statistical physics case, the vectors ${\bf w}$ are typically low dimensional, whereas we are interested in very high dimensional parameter vectors. Modern machine learning applications often deal with parameter vectors with a dimensionality in the $10^6 - 10^9$ range. Note that although not relevant for real physical systems, interaction energies with such high dimensional vectors are still of theoretical interest to physicists since mean field analysis can become exact in this limit. 

{\it More general loss functions:} Convergence proofs in the relevant literature are often presented for more general loss functions than considered in this manuscript. In the interpolating limit, the loss function reduces to a quadratic form near the interpolating minimum and the convergence analysis can be performed by bounding the largest singular value of the linear operator governing the parameter iteration. More general loss functions treated in the literature are coupled with smoothness and strong convexity constraints. However, these constraints in turn imply that the loss functions are bounded above and below by a quadratic form, at least near the optimum. It should therefore be possible to reformulate the results presented here into a more general context coupled with smoothness and strong convexity constraints. It is not clear that fundamental insights will be gained by such a reformulation, which would introduce more notational complexity. So we constrain ourselves to the quadratic loss.   

\subsection{Fast convergence of SGD in the interpolating limit}

The starting point for this work is the recent work by Belkin {\it et al}\cite{ma2017power} presenting an analysis of the fast convergence of the SGD algorithm in the interpolating limit. We briefly recount some of the results of this paper and also present a modified SGD algorithm that allows for the derivation of exact formulae for the convergence rate. This helps us understand the efficiency of distributed computation for the problem at hand. 

Consider the standard supervised learning setup with a data set consisting of labelled pairs $(y_i, {\bf x}_i)$, $i=1..n$. The task is to learn a parametrized function $f({\bf x},{\bf w})$ chosen from a suitable function class, by minimizing the empirical risk ${\bf w}^*=argmin L({\bf w})$, corresponding to a loss function 

\begin{align} 
L({\bf w}) &= \frac{1}{n}\Sigma_i l_i({\bf w})\\
l_i({\bf w}) &= l(y_i,f({\bf x}_i,{\bf w}))
\end{align} 

{\it Quadratic loss in the interpolating limit:} The interpolating limit is defined by the conditions $l_i({\bf w}^*)=0$, ie the loss is zero at each sample point (ie the interpolating function passes through each data point). In this limit $l(y_i,f({\bf x}_i,{\bf w}))$ is close to zero if $|{\bf w}^*-{\bf w}|$ is small. We will assume it corresponds to a quadratic loss in the neighborhood of ${\bf w}^*$, and that the function $f$ is differentiable at ${\bf w}^*$. 

Under these conditions $l_i({\bf w}) \approx ({\bf X}_i\cdot({\bf w}^*-{\bf w}))^2$ where ${\bf X}_i$ is $\nabla_x f({\bf x}_i,{\bf w}^*)$. Suitably redefining variables as ${\bf X}_i\cdot{\bf w}^* =\tilde{y}_i$ and ${\bf X}_i=\tilde{\bf x}_i$, we see that we are left with a loss function corresponding to a linear model, $ l_i({\bf w}) = (\tilde{y}_i -\tilde{{\bf x}}_i\cdot {\bf w})^2 = (\tilde{{\bf x}}_i\cdot {\bf \delta w})^2$, where $ {\bf \delta w} = {\bf w} - {\bf w}^*$. In the following we will drop the tildes for notational simplicity. We are now effectively analyzing linear regression, and we denote the dimensionality of ${\bf x}$ and ${\bf w}$ by d. 

{\it Overfitting:} To be in the interpolation regime one generally needs to overfit, ie $d\geq n$. Note that in this case the Hessian has a number of zero singular values, corresponding to a null space about which the data is not informative. The ERM procedure will not reduce error in this null space, so our attention will be confined to the range of H corresponding to its {\it nonzero} singular values. In the linear regime the null space is left invariant. For notational simplicity the vectors ${\bf w}_i$ denote only the projections orthogonal to the null space. The projection parallel to the null space is simply left invariant by the iterative procedures below. 

With the above setup it is easy to verify that the corresponding GD algorithm is given by ($\eta$ is the learning rate): 

\begin{eqnarray}
{\bf \delta w}_{t+1} &=& (1-\eta H ){\bf \delta w}_{t}\\
H &=& \frac{1}{n}\Sigma_{i=1}^n x_i^2 P_i\\ 
x_i &=& |{\bf x}_i|_2 ~~~~~~~~~~
P_i = \hat{\bf x}_i \hat{\bf x}_i^T
\end{eqnarray}

{\it Notational Choices:} We have made some notational choices to simplify the following considerations and written the Hessian H (equivalently the sample covariance of the vectors ${\bf x}$) in terms of a sum of projection operators $P_i$ corresponding to the individual data points. Note that $P_i^2=P_i$. If the vectors  ${\bf x}_i$ are orthogonal then $P_i P_j = 0$ when the indices are unequal. In the general non-orthogonal case the $P_i$ do not commute. 

{\it Stochastic formulation:} We now introduce iteration-dependent stochastic binary variables $\sigma_i(t) \in \{0, 1\}$ where the variables will be chosen i.i.d from a Bernoulli distribution, with $E(\sigma_i) = \frac{m}{n}$. The idea is that $\sigma_i(t) = 1$ iff the sample $i$ is picked in the minibatch used in the $t^{th}$ time step. Note also that the GD case is recovered for $m=n$ since each sample is picked with probability 1. 

{\it Difference from standard procedure - stochastic minibatch size:} This stochastic formulation is slightly different from the usual setting, where mini batches are picked with fixed size m. In the formulation presented here, the batch sizes fluctuate from one iteration to another, with an {\it average} size of $E[\Sigma_i \sigma_i(t)] = m$. It is easy to treat the fixed batch size case using the same formulation, as long as one keeps track of the correlations introduced between different $\sigma_i$ at a given iteration due to the constraint of the fixed batch size, but we will not present the corresponding formulae here. 

{\it Samples drawn without replacement:} Note that the minibatch sampling procedure used by Belkin {\it et al}\cite{ma2017power} is {\it with replacement}, which is slightly different from the setting here or in standard SGD. The reason we introduce this variant of the SGD algorithm is that the randomness has been made explicit as an uncorrelated binary process associated with each sample, which makes the analysis simpler. 

{\it Specific SGD algorithm analyzed in this paper:} With the above notation we define an SGD algorithm (note the stochastically variable batch size) by the iteration 

\begin{eqnarray}
{\bf \delta w}_{t+1} &=& M(t){\bf \delta w}_{t}\\
M(t) &=& 1-\frac{\eta}{m}\Sigma_{i=1}^n x_i^2 \sigma_i(t) P_i 
\end{eqnarray}

Note that $\sigma_i(t)$ is uncorrelated with ${\bf \delta w}_{t}$. Therefore $E|{\bf \delta w}_{t+1}|^2$ can be computed as $E[{\bf \delta w}_{t}^{\dagger} E[M^{\dagger}(t)M(t)]{\bf \delta w}_{t}]$, where the expectation is over the stochastic processes $\sigma_i(t)$.  To analyze convergence one needs to bound the largest singular value of the matrix $E[M^{\dagger}(t)M(t)]$ and optimize it with respect to the learning rate $\eta$. 

{\it Orthogonal sample vectors:} Let us first consider the case where the sample vectors are orthogonal (equivalently $P_i P_j = 0$ for $i \neq j$). Then, noting that $E[\sigma_i(t)]=E[\sigma_i^2(t)]=m/n$, we have

\begin{equation}
E[M^{\dagger}(t)M(t)] = 1-2\eta H + \frac{n}{m}\eta^2H^2
\end{equation}

Thus in the orthogonal case the eigenvalues of $E[M^{\dagger}(t)M(t)]$ are given by $1-2\frac{\eta}{n}x_i^2 + \frac{\eta^2}{mn}x_i^4$. To obtain the best bound for the decay rate one has to maximize this expression over $i$, then minimize that result over $\eta$. Consider the case $x_i=1$, ie {\it orthonormal} sample vectors. Then the eigenvalues are all equal and are given by $1-2\frac{\eta}{n} + \frac{\eta^2}{mn}$. The minimum value is obtained for $\eta=m$, and is given by $g(m) = (1-\frac{m}{n})$. If the $x_i$ are not all equal to 1, we get $g(m)=(1-c\frac{m}{n})$ where $c=(GM)^2/(AM)^2$, with $GM$ and $AM$ being the geometric and arithmetic means of $x_{min}^2$ and $x_{max}^2$. Thus $g(m)$ is less than 1 for any $m$ and this shows exponential convergence to zero error with iteration number. 

{\it Gain from paralellization:} The number of iterations required to achieve (on average) a relative error of $\epsilon$ is given by $g(m)^t=\epsilon$ ie $t_{\epsilon}=\log(1/\epsilon)/\log(1/g(m))$, whereas during that same time a computational cost of $mdt_{\epsilon} = md\log(1/\epsilon)/\log(1/g(m))$ is incurred. The total computation cost to achieve a fixed total error depends on the batch size as $m/\log(n/(n-cm))$. This cost decreases as $m$ increases, so that bigger batch sizes will produce the same error at a lower computational cost, indicating that the problem will continue to benefit from data parallelism as $m$ increases. 

However, the situation is different if the data vectors have strong correlations, and in particular the Hessian matrix has a few large eigenvalues that dominates its trace. In this case, Belkin {\it et al}\cite{ma2017power} show that the gain from parallelism is limited, and that the parallelism gains saturate when m reaches a value $m^*$ given by $Tr(H)/\lambda_{max}(H)$. 

{\it Non-orthogonal sample vectors:} Next we consider the more general case where ${\bf x}_i$ are not orthogonal but are normalized ($x_i^2=1$). Noting that $E[\sigma_i \sigma_j]=(\frac{m}{n})^2 + (1-\delta_{ij})(1-\frac{m}{n})$ it is easy to show that $E[M^{\dagger}(t)M(t)] = (1-\eta H)^2 + \frac{\eta^2}{m}(1-\frac{m}{n}) H$. As expected setting $m=n$ we recover the GD matrix. the eigenvalues of $E[M^{\dagger}(t)M(t)]$  are then given by

\begin{equation}
g(m,\eta,\lambda_i) = (1-\eta \lambda_i)^2 + \frac{\eta^2\lambda_i }{m}(1-\frac{m}{n})
\end{equation}

where $\lambda_i$ are the eigenvalues of the sample covariance or Hessian $H$. The bound on the decay rate is given by $min_\eta \max_i g(m,\eta,\lambda_i)$. Since $g$ is quadratic in $\lambda$ the maximum over $\lambda_i$ is achieved at either $\lambda_1 = \lambda_{max}(H)$ or $\lambda_n=\lambda_{min}(H)$. For fixed $\lambda$ the dependence on $\eta$ is also quadratic. If one plots $g(m,\eta,\lambda)$ vs $\eta$ for $\lambda=\lambda_n$ and $\lambda=\lambda_1$ one obtains two intersecting parabolas. The minimum tracks one parabola and then the other, with the overall minimum occurring when $g(m,\eta,\lambda_1)=g(m,\eta,\lambda_n)$. Solving this equation one obtains (assuming $\lambda_1>\lambda_n$)

\begin{eqnarray}
\eta^*(m) &=& \frac{2}{\lambda_1+\lambda_n+\frac{1}{m}-\frac{1}{n}}\\
g^*(m)&=& 1-\frac{4 \lambda_1 \lambda_n}{(\lambda_1+\lambda_n+\frac{1}{m}-\frac{1}{n})^2}
\end{eqnarray}

For $m=n$ one obtains the GD result 

\begin{equation}
g^*(n) = (\frac{\lambda_1-\lambda_n}{\lambda_1+\lambda_n})^2 = (1-\frac{2}{C+1})^2
\end{equation}

where $C=\frac{\lambda_1}{\lambda_n}$ is the condition number of $H$. If $\lambda_{max} \sim 1$, as would be the case when there are strong correlations, then $g^*(m)$ approaches $g^*(n)$ fairly quickly. For some parameter choices the total computation cost $mdt_{\epsilon} = md\log(1/\epsilon)/\log(1/g(m))$ shows a minimum for a value of $m>1$ but close to 1. However if $\lambda_1\sim\frac{1}{n}$ (this is the case for the orthogonal matrix) then $g^*(m)$ approaches $g^*(n)$ more slowly. Thus the optimal choice of $m$ is dependent on the degree of correlations between then normalized sample vectors. 

{\it Remark:} Note that in contrast with the original analysis presented in Belkin {\it et al}\cite{ma2017power} we are able to derive exact formulae for the optimal learning rate $\eta^*(m)$ and convergence rate $g^*(m)$ rather than bounds. Derivation of these exact formulae rely on the variant of the SGD introduced in this paper, in which each sample is chosen with a fixed probability at each time step and the minibatch sizes vary stochastically. 

\section{Distributed GD with Elastic Penalty on a Graph in the Interpolating Limit}

These considerations demonstrate that (i) SGD shows rapid convergence in the interpolation regime and that (ii) data parallelism should be computationally beneficial if the sample vectors are not strongly correlated. Note that data parallelism is not dictated by computational cost alone - it may be practically impossible to store data locally at a central compute node, and one has to also consider the communication costs of centralized parallel computation using a parameter server. 

{\it Necessity of Distributed SGD - Communication cost of a central parameter server:} Even for data-parallel implementations of SGD, centralized communication to a parameter server may cause a problem. In the extreme case, where each compute node has one data vector, communicating all $n$ parameter vectors to a central server after gradient updates, requires a communication link with bandwidth $n*d$. With $n,d$ both $\sim 10^6-10^9$, this may be impossible to provide. With these motivations we proceed to study the fully distributed case (without a parameter server), where an individual compute node only communicates locally with the set of nodes connected to it. For simplicity we consider the case of a fixed, connected graph, although similar results should continue to hold on a fluctuating graph topology as long as the fluctuations still permit diffusion of signals. 

{\it Problem setup:} We assume there are $n$ compute nodes, each with a single data vector, and a node-specific parameter vector. Define the penalized loss function as in Eq.(1), ie $L({\bf w}_1,...,{\bf w}_n) = \Sigma_i l_i({\bf w}_i) + \mu \sum_{<i,j>}|{\bf w}_i-{\bf w}_j|^2$. The set of edges $<i,j>$ specify the communication graph between the nodes. We do not constrain this graph beyond the requirement that it should be connected. 

{\it Simplification in the interpolation limit:} Generally, the penalty term will {\it not} be minimized by a set of ${\bf w}_i$ that also minimize an un-penalized loss function with this form. However, the {\it interpolating limit} is special since there exists a vector ${\bf w}^*$ that minimizes each $l_i({\bf w})$, that minimum value being zero ie $l_i({\bf w}^*)=0$ for all i. Clearly, the penalty term also equals zero if ${\bf w}_i={\bf w}^*$. Since all terms in the sum are non-negative, it can be seen that an interpolating minimum of the loss is {\it simultaneously} a minimum of the penalized loss. This considerably simplifies things. Note again that we will ignore the presence of zero modes as the GD dynamics will leave a null subspace unchanged. 

{\it Algorithm:} The distributed GD algorithm is given by 

\begin{equation}
\label{DSGD}
{\bf \delta w}_i^{t+1} = (I-\eta x_i^2 P_i){\bf \delta w}_i^{t} + \mu \sum_j L_{ij} {\bf \delta w}_j^{t}
\end{equation}

{\it Notation:} Here $L$ is the Graph Laplacian defined by the quadratic form ${\bf W}^{\dagger} L {\bf W} = -\Sigma_{<i,j>} |{\bf w}_i-{\bf w}_j|^2$. We have defined a concatenated vector ${\bf W} = [{\bf w}_1; {\bf w}_2; ..; {\bf w}_n]$. 

{\it Proof outline:} To prove exponential convergence of this distributed GD algorithm, one only needs to show that the self-adjoint linear operator $I-\eta {\bf H} +L$ governing the dynamics has its largest eigenvalue less than 1. 

{\it Equivalently}, we need to show that the smallest eigenvalue $\sigma_{min}$ of the $nd \times nd$ matrix $\eta {\bf H} -L$ is greater than zero, where $\eta {\bf H}$ has diagonal blocks given by $x_i^2 P_i$. Note that the largest eigenvalue of the evolution operator is $1-\sigma_{min}$. 

It is convenient to start with the expression   
$\sigma_{min} = min [{\bf \hat{W}}^{\dagger} (\eta {\bf H} -L) {\bf \hat{W}} ]$, with the minimum being taken over unit vectors $ |{\bf \hat{W}}|=1$. The proof follows by expanding out the quadratic form: 

\begin{equation}
\label{sig}
\sigma_{min} = min_{| \bf \hat{W} |=1}\bigl[ \eta \Sigma_{i=1}^n ({\bf x}_i^{\dagger} \hat{\bf w}_i)^2 + \mu \sum_{<i,j>} |\hat{\bf w}_i-\hat{\bf w}_j|^2 \bigr]
\end{equation}

We have to demonstrate that $\sigma_{min} >0$. Note that the argument being minimized is a sum over squares, so for the sum to be zero, each individual term must be simultaneously zero. However it is easy to show that this is impossible. Writing $\sigma_{min}=min[Term1+Term2]$ with $Term2$ consisting of the Laplacian penalty, we will show that $Term2=0 \implies Term1 >0$. 

Since the communication graph is connected, the Laplacian penalty $Term2$ will only vanish if $\hat{\bf w}_i$ are all equal, 
$\hat{\bf w}_i = \frac{1}{\sqrt{n}}\hat{\bf w}$ for all $i$. The vectors $\hat{\bf w}$ are normalized in the d-dimensional sample space, and the extra normalization factor ensures that  $|{\bf \hat{W}}|=1$). 

Plugging this choice of  ${\bf \hat{W}}$ into $Term1$ we get $Term1=\frac{\eta}{n} \hat{\bf w}^{\dagger} H \hat{\bf w}$. Recall that we are only considering the subspace corresponding to the non-zero eigenvalues of $H$ (the dynamics leaves the null space of $H$ invariant). This implies that  $Term1=\frac{\eta}{n}\hat{\bf w}^{\dagger} H \hat{\bf w} \geq \frac{\eta}{n}\lambda_{min}(H) >0$ where $\lambda_{min}(H)$ is the smallest nonzero eigenvalue of $H$. Thus, $Term2=0 \implies Term1>0$.  

Thus, there is no choice of ${\bf \hat{W}}$ for which both  $Term1=0$ and $Term2=0$. It follow that $\sigma_{min}>0$. This argument does not provide an explicit estimate of $\sigma_{min}$, but the argument above shows that that $\sigma_{min}<\frac{\eta}{n}\lambda_{min}(H)$. Thus the largest eigenvalue of the evolution operator in Eq.(\ref{DSGD}) is (strictly) between 1 and $1-\frac{\eta}{n}\lambda_{min}(H)$. 

{\it Convergence rate estimate:} This concludes the proof that the error in the distributed GD algorithm shrinks exponentially to zero in the interpolating limit, with a rate 

\begin{equation}
1-\frac{\eta}{n}\lambda_{min}(H)<R<1
\end{equation}

If the penalty term is large ($\mu\rightarrow \infty)$ then we can expect that the first term in the quadratic form will dominate and $R\rightarrow 1-\frac{\eta}{n}\lambda_{min}(H)$. 

\section{Conclusion and Discussion}

In this manuscript we have exploited the simplicities arising in the interpolating limit for function learning, to analyze the convergence of stochastic as well as distributed gradient descent close to an interpolating minimum of the loss function. While the analysis is simple and is based on linear regression using a least squared loss, we expect the conclusions to hold in qualitative terms for more general loss functions, with suitable smoothness and strong convexity constraints near the interpolating minimum. 

\begin{itemize}
\item We have introduced a variant of SGD in which data samples are picked using i.i.d Bernoulli processes. In contrast with the standard SGD algorithm, for this variant the minibatch sizes fluctuate stochastically from one iteration to the next with a Binomial distribution. This simplifies theoretical analysis and allows us to explicitly compute the optimal learning rate $\eta^*(m)$ and corresponding convergence rate $g^*(m)$. This approach may have more general theoretical utility than shown here. The analysis also shows the importance of the correlation structure of the input vectors in determining the efficiency of SGD. 

\item The empirical efficiency of the SGD algorithm for small minibatch sizes points to the presence of strong correlations in real life data sets - even though the input dimensions are nominally very large, it is possible that the effective dimensionality is still modest. 

\item We have presented and analyzed a distributed Gradient Descent algorithm also in the interpolating limit, with each compute node holding only one data sample. We have shown that a Graph Laplacian-penalized distributed loss function adequately couples the nodes to drive the system to an interpolating minimum, with error exponentially decreasing with iterations (for finite sized connected graphs). 
\end{itemize}

We have not specified the communication graph beyond the requirement that it is connected. In order to minimize communication costs, this graph should be adequately sparse. Design of an optimal graph for the distributed GD algorithm for a fixed communication cost is an interesting problem which we have not pursued here, but it is a direction for future research. 

Stochastic and asynchronous variants of the algorithm presented should be interesting to analyze (eg where the gradient update step is decoupled from the diffusion step). 

One possibility not developed here, is to run the distributed GD algorithm in Eq.(\ref{DSGD}) in the {\it under-parametrized} regime, with the connectivity graph between data nodes determined by similarities in the sample vectors. In this case, one will generally not be in the interpolating limit and the individual losses will not all be reduced to zero. However the Laplacian penalty term would enforce local smoothness of the parameter vector on the connectivity graph. This would amount to a form of local linear regression.  

\section*{Acknowledgment}

PPM Thanks Misha Belkin and Saikat Chatterjee for extensive discussions. Support from the Crick-Clay Professorship at Cold Spring Harbor Laboratory and the H N Mahabala Chair Professorship at IIT Madras is gratefully acknowledged. 

\bibliography{EUSIPCO2018}
\bibliographystyle{ieeetr}

\end{document}